\documentclass[%
conference]{IEEEtran}
\IEEEoverridecommandlockouts
\usepackage{amsmath,amssymb,amsfonts}
\usepackage{algorithmic}
\usepackage{graphicx}
\usepackage{textcomp}
\usepackage{xcolor}

\usepackage{biblatex}
\bibliography{references}
\AtEveryBibitem{
    \clearfield{urlyear}%
    \clearfield{urlmonth}%
    \clearfield{note}%
    \clearfield{url}%
    \clearfield{eprint}%
    \clearfield{issn}%
}

\def\BibTeX{{\rm B\kern-.05em{\sc i\kern-.025em b}\kern-.08em
    T\kern-.1667em\lower.7ex\hbox{E}\kern-.125emX}}

\begin{document}

\title{Benchmarking Robustness of Deep Reinforcement Learning approaches to Online Portfolio Management
\thanks{This work benefited from the support of the Chair "Artificial intelligence applied to credit card fraud detection and trading" led by CENTRALESUPELEC and sponsored by LUSIS. This work was performed using HPC resources from the "Mésocentre" computing center of CentraleSupélec, École Normale Supérieure Paris-Saclay and Université Paris-Saclay supported by CNRS and Région Île-de-France (https://mesocentre.universite-paris-scalay.fr/).}
}

\author{\IEEEauthorblockN{Marc Velay\IEEEauthorrefmark{1},
Bich-Liên Doan\IEEEauthorrefmark{1}, 
Arpad Rimmel\IEEEauthorrefmark{1} , 
Fabrice Popineau\IEEEauthorrefmark{1} 
and Fabrice Daniel\IEEEauthorrefmark{2}}
\IEEEauthorblockA{\IEEEauthorrefmark{1}
Université Paris-Saclay, CNRS, CentraleSupélec\\
Laboratoire Interdisciplinaire des Sciences du Numérique\\
91190 Gif-sur-Yvette, France\\
Email: firstname.lastname@centralesupelec.fr}
\IEEEauthorblockA{\IEEEauthorrefmark{2}
LUSIS\\
5 cité Rougemont\\
75002 Paris, France\\
Email: fabrice.daniel@lusis.fr}}

\maketitle

\begin{abstract}
Deep Reinforcement Learning (DRL) approaches to Online Portfolio Selection (OLPS) have grown in popularity in recent years. The sensitive nature of training Reinforcement Learning agents implies a need for extensive efforts in market representation, behavior objectives, and training processes, which have often been lacking in previous works. We propose a training and evaluation process to assess the performance of classical DRL algorithms for portfolio management. We found that most DRL algorithms were not robust, with strategies generalizing poorly and degrading quickly during backtesting.
\end{abstract}



\section{Introduction}

Improvements in Reinforcement Learning (RL), specifically Deep Reinforcement Learning (DRL) have contributed to new state-of-the-art performances in many fields from robotics \cite{lillicrap_continuous_2016}, autonomous vehicles \cite{wang_deep_2019} and games \cite{mnih_human-level_2015}. These improvements have also led to a broad application to fields previously dominated by heuristic approaches. Online Portfolio Selection (OLPS) has seen a large increase in the popularity of DRL methods, leading to seemingly conclusive and positive results. OLPS is a dynamic management of financial assets, where we have the opportunity to redistribute funds across assets regularly, a sequential decision-making problem. This allows the managing algorithm to adapt to market changes with the aim of outperforming index performance. 
  
However, evaluations of existing works often rely on limited metrics and protocols, which may not be suitable for portfolio management. Many works cannot be reproduced due to data unavailability or lack of experimental information. OLPS is unlike many other RL problems due to its high uncertainty and non-stationary nature. The financial market, used as the RL agent's environment, has a disproportionate impact on the proposed reward functions and  lack of predictability. Furthermore, RL algorithms are very sensitive to hyperparameter selection and initialization, which require additional evaluations and consideration. Many published results often contain single-initialization results, which may misrepresent the capabilities of an approach and yield poorer performance if deployed. Furthermore, traditional metrics only indicate the performance of an algorithm during its conception phase, close to its training data. We have found that agents tend to overfit, picking the same assets regardless of market variations, which reflect favorably in traditional metrics. However, when the market evolves, the static behavior degrades. Evaluating the robustness of algorithms and their capacity to adapt to uncertainty and out-of-distribution data gives greater insight into their training quality and generalization capabilities. 

The aim of this work is to provide a standardized comparison process to assess portfolio management algorithms. The process provides reproducible results on the performance of the management agents. Reinforcement Learning is a complex task with many components sensitive to tuning and design choices. The proposed training and evaluation setup measures the robustness and generalization capabilities of our benchmarked algorithms. We rely on public data that are freely available and open-source implementations of DRL algorithms to obtain transparent comparisons of popular approaches to OLPS. We focus on evaluating the training quality of our agents and their robustness to never-seen market conditions. To the best of our knowledge, there is no large comparison of approaches in this domain, with multi-dimensional evaluation of components' contribution to performance.

\section{Related Works}

Previous works have often focused on single Deep Reinforcement Learning approaches, generally improving one aspect of the problem. In this section, we compare different methods that have been previously presented. We focus on the learning algorithms used, the ways to represent the environment, how they modify the portfolio positions, and how they influence the learned behavior through rewards. These related works provide the different components which we evaluated in our experiment. 

\subsection{Learning Algorithms for OLPS}

\begin{table}[htbp]
\caption{Popular Learning Algorithms\label{popular-rl}}
\begin{center}
\begin{tabular}{|c|c|c|c|}
\hline
 DDPG & PPO & A2C & SAC \\ 
 \hline
\cite{benhamou_bridging_2020,durall_asset_2022,jiang_deep_2017,li_optimistic_2019,liang_adversarial_2018,ye_reinforcement-learning_2020,zhang_cost-sensitive_2020} & \cite{durall_asset_2022,liang_adversarial_2018,ye_reinforcement-learning_2020} & \cite{durall_asset_2022} & \cite{durall_asset_2022} \\
\hline
\end{tabular}
\end{center}
\end{table}

The most popular learning algorithm from Table ~\ref{popular-rl} is Deep Deterministic Policy Gradient (DDPG), an iteration upon the Policy Gradient algorithm, which is an efficient way to include continuous states and actions. We found works which made use of more recent algorithms that also improve upon Policy Gradients, such as Proximal Policy Optimization (PPO), Advantage Actor-Critic (A2C), and Soft Actor-Critic (SAC). Some authors modified the algorithms to improve generalization capabilities through data augmentation \cite{benhamou_bridging_2020,li_optimistic_2019,liang_adversarial_2018}, resulting in some performance improvements. Others proposed policy ensembles \cite{yang_deep_2020}, with the best-performing policy on recent historic data controlling the portfolio for a given time. 

PPO is the only on-policy algorithm, learning from its recent experiences. DDPG, A2C, and SAC are off-policy algorithms exploring many paths in the environment for each learning iteration. This approach can avoid staying in local optima by having a broader spectrum of experiences. However, this keeps older experiences relevant during learning. This is efficient for problems with large spaces to explore and data availability to support such a process, where transitions are infrequently replayed. OLPS has limited data, with roughly 2500 points for 20 years of history. This is a limiting factor that may have motivated some authors to augment their data. However, this can also be mitigated by sampling subsets of assets, where each set represents a new distinct environment.  
  
PPO and SAC include mechanisms to increase gradient step stability, such as entropy regularization or constraining step size to the neighborhood of previous weights. This may be useful for high-uncertainty environments. DDPG and A2C do not provide such mechanisms, however they provide the ability to evaluate an actor through multiple critics.

\subsection{Market Representations}

\begin{table}[htbp]
\caption{Popular Market Representations\label{popular-market-representations}}
\begin{center}
\begin{tabular}{|c|c|c|c|}
\hline
 Current Prices & Sliding Window & Sparse Window & Context \\ 
 \hline
\cite{durall_asset_2022,li_optimistic_2019,yang_deep_2020} & \cite{jiang_deep_2017,liang_adversarial_2018} & \cite{benhamou_bridging_2020,pigorsch_high-dimensional_2021} & \cite{benhamou_bridging_2020,pigorsch_high-dimensional_2021,ye_reinforcement-learning_2020,zhang_cost-sensitive_2020} \\
\hline
\end{tabular}
\end{center}
\end{table}

Each representation of the market in Table ~\ref{popular-market-representations} must contain enough information for an agent to select the best action. Based on MDPs, this should only contain information about the current state as they are independent. This theoretically sound approach was tried in previous works and may include financial indicators, aggregates of past prices, current asset allocations, or asset correlations. However, modeling complex time series often requires past data points for more accuracy, which can be done using sliding windows of past prices, which may be considered unique states. This was improved by using sparse sliding windows, drastically reducing the state space.

Some previous works argue that the financial market may be a Partially Observable MDP, requiring additional contextual information to accurately select actions. Frequent uses include NLP encodings or historical price statistics. A rare approach is Technical Analysis Indicators used by human traders.

\subsection{Management Rewards}

\begin{table}[htbp]
\caption{Popular Rewards\label{popular-rewards}}
\begin{center}
\begin{tabular}{|c|c|c|c|}
\hline
 Daily Returns & Episodic Returns & Rate of Return & Risk \\ 
 \hline
\cite{durall_asset_2022,li_optimistic_2019,liang_adversarial_2018,pigorsch_high-dimensional_2021,yang_deep_2020,ye_reinforcement-learning_2020} & \cite{benhamou_bridging_2020} & \cite{jiang_deep_2017} & \cite{zhang_cost-sensitive_2020} \\
\hline
\end{tabular}
\end{center}
\end{table}

The rewards in in Table ~\ref{popular-rewards} control the desired behavior of agents. Different metrics can favor risk-taking, portfolio turnover, or reach a trade-off between the two. The most popular approach employed by previous works consists of measuring the net difference in portfolio value between two days. This difference can be between consecutive days or a total difference over the trading period. However, this approach is nonstationary, as large changes in valuation between the start and end of a trading period can impact their value, leading to perceived changes in rewards while they are proportionally similar. To remedy this caveat, some works propose using a rate of returns or an average daily log return. An alternative uses a composite function that rewards gains but constrains the turnover through regularization.

\subsection{Limitations}

Most of the previously cited works lack one or more measures to evaluate their algorithms fully. Some publications use private data, which may never be reproduced due to their nature. Others lack information about selected assets or the periods during which they have evaluated performance. The use of public financial data, with clearly defined training and backtesting processes, may remedy this caveat and allow others to reproduce our results.

All works evaluate their algorithms with classic financial metrics, such as net returns, risk metrics, including Sharpe and Sortino ratios or Maximum DrawDown. While these do well to evaluate portfolio management, we have found that overfitting agents predicting the same allocation weights regardless of selected assets and market changes had good performance measures in validation but would quickly become obsolete if deployed. Worse, many previous works only share a single result for their approaches, lacking confidence measures. Combined with the difficulties of training DRL agents, the same algorithm implementation with the same data may yield completely different results if initialized with different seeds. While the published results may be encouraging, they may not be representative of their validity for future periods. Training multiple agents, with different initializations, for each approach would allow the computation of confidence metrics to validate the results.

Most works are compared to classic portfolio optimization algorithms, such as Mean-Variance Optimization, while few compare to other DRL approaches. Combined with the lack of reproducibility from the first point, we cannot determine why we should choose their approach compared to other solutions. This motivates us to include a broad range of methods in this work to determine the preferred use of learning algorithms, market representations, and behavior objectives.

\section{Deep Reinforcement Learning for Online Portfolio Selection}

Based on previous works, we define our OLPS environment such that an agent allocates wealth across N assets over T time steps. The actions directly impact the weights of the assets in the portfolio, such that each asset corresponds to an action dimension. We add the possibility for the agent to keep liquidity to reduce portfolio volatility. 
$$a_t=\{a \in \mathbb{R}^{N+1};\ \sum_{i=0}^{N} a_i=1, a_i \in [0,1]\}$$
For each time step, after rebalancing the portfolio, we observe a transition, composed of a reward and a new state. The aim of training agents is to find the sequence of actions that maximizes the rewards obtained during an episode. 

We select a set of four rewards to optimize the behavior of the algorithms based on previous works. \textit{Daily net returns} are frequently used and correspond most to the MDP framework. Yet, it is complex to attribute a reward to each action, as it is influenced by previous allocations and price changes, leading to delayed payoffs. Episodic rewards evaluate management over T time steps, smoothing responsibility from individual actions to sequential decisions. \textit{Total Net Returns} measures the total wealth changes over the T periods. \textit{Episodic Rate of Returns} is numerically more interesting, being stationary, with a value range closer to RL recommendations. Finally, the \textit{Sortino ratio}, described in the experiments section, is a measure of risk, which may lead to more careful management. We aimed to compare the influence of each reward function on management capabilities.

Agents must learn the optimal actions for given states to maximize rewards. The choice of market representations influences the quality of state-action-reward mapping. Representations containing more information add complexity through higher state dimensions. Using only market data, composed of Open, High, Low, and Adjusted Closing prices, and Volume, we select four market representations. The \textit{Markovian Representation} is the default, using only the five normalized OHLCV values for each asset for a given time t. This results in a tensor of one dimension $\mathbb{R}^{N * 5}$. To add more information about the recent fluctuations, we use a continuous \textit{Sliding Window Representation}, containing the normalized OHLCV values of the last month. This state is composed of a two-dimensional tensor of shape $\mathbb{R}^{21 \times N * 5}$. A solution to reduce the substantial state size of the windowed representation is to use a sparse sliding window, as individual distant prices have less influence on coming variations. The \textit{Lagged Representation} only uses data from the past five open days and a day 2-3-4 weeks ago, reducing the state dimensions to $\mathbb{R}^{8 \times N * 5}$. Finally, human traders have used technical analysis for the past decades to better understand the market. The \textit{Indicators Representation} uses raw OHLCV values to compute a synthesis of past market conditions. We include the change since the last time step, MACD, Bollinger bands bounds, RSI, CCI, DX, and a smoothed 30 days moving average of closing prices for a state tensor of dimension $\mathbb{R}^{N * 8}$.

Finally, to learn the state-action-reward mapping, we rely on popular DRL algorithms. We select SAC, PPO, DDPG, and A2C based on their previously mentioned differences and benefits.

\section{Experiments}

In this section, we present market data acquisition and processing, algorithm details and training, and performance measures.

\subsection{Data Processing}

We focused on S\&P500 constituent stocks at the date of January 2023, for the period from January 2010 to December 2022. We reserved the years 2021-2022 for backtesting and used the remaining 11 years for training. We analyzed 500 assets, from which we composed a 20-asset portfolio. During training, 20 stocks are randomly subsampled. Due to the evolving nature of the market, not all assets existed at the start of the historical data. We excluded those missing from being sampled, resulting in a growing stock pool as dates approach the present. The training period was randomly split into non-overlapping training and validation sets of 60 days. We accounted for the range of historical data required for some market representations, such as sliding windows, to avoid leaking validation data into training.

For backtesting, we manually selected 20 assets from diverse industries with moderate to high returns for the period. This guarantees that every evaluation was done in the same universe, while limiting performance skewing from selecting only top-performing assets. 

Instead of using raw Open, High, Low, and Close prices, we normalized them to obtain stationary values. The values we used in most market representations, excluding indicators, is $x_t=\frac{p_t}{p_{t-1}}-1$, where p is the price. The starting value of the portfolio was 100000USD.

\subsection{Training agents}

Reinforcement Learning algorithms are very sensitive to both initialization and hyperparameter selection. The first step is to find a configuration that converges after a given number of iterations. For each algorithm, we ran 100 trials with Optuna TPE sweeper, which randomly samples sets of parameters around the previous best runs. The samples vary for each approach based on the RL algorithm used. However, the volume of parameters to set for the environment, neural networks, and learning algorithms is very large. From experience and domain knowledge, we selected parameters with fixed, reasonable values, that were excluded from sweeps. This is also motivated by the substantial amount of compute time required to train a single trial in a sweep, ranging between 2 and 5 hours on a CPU with 40 cores. The market representations were set based on reasonable values, such as past month sliding windows and most frequently used financial indicators. This approach is motivated by the less sensitive nature of the representations to small changes.

During the hyperparameter sweep, we trained an agent on the training set and evaluated it on the validation sets. Their starting dates and lengths were shared across all trials to avoid bias through market conditions. The result is a set of hyperparameters for which training converges for a given algorithm combination. We used this set to train multiple agents for the same algorithm using different initialization seeds. This training phase was longer than the trials, with no early stops. The best checkpoint on the validation set was selected as representative. Limited by compute time, we trained five agents for each algorithm, using different initialization seeds \cite{henderson_deep_2018}. The end result is 320 trained agents to backtest.

\subsection{Backtesting Evaluation}

We backtest all our algorithms on the two-year period of 2021-2022. We manually selected 20 assets from different sectors (see table~\ref{assets-backtest}). All agents, heuristic or DRL-based, were evaluated on the same period. Some representations required historical data, which overlaps the training data. 

\begin{table}[htbp]
\caption{Assets used in backtest\label{assets-backtest}}
\begin{center}
\begin{tabular}{|c c c c c|} 
 \hline
 Tech & Healthcare & Industry & Finance & Energy \\ [0.5ex] 
 \hline\hline
 AAPL & UNH & ALLE & JPM & XOM \\ 
 \hline
 MSFT & JNJ & AME & BAC & CVX \\
 \hline
 GOOG & PFE & BA & WFC & NEE \\
 \hline
 AMZN & ABBV & CAT & MS & COP \\
 \hline
\end{tabular}
\end{center}
\end{table}

Using the logged daily actions, we computed the performance and robustness metrics. For performance, we use traditional finance metrics. The first is the portfolio returns, defined as $PR=\text{Value}_{t} - \text{Value}_{t-n}$, with n the comparison period. Specifically, $n=1$ the daily returns and $n=T$, T being the length of the backtesting period. Portfolio Returns are used as the foundation of most other metrics. From net returns, we obtain the rate of returns, a metric independent of the initial funds, defined as $RoR=\frac{\text{Value}_{t}-\text{Value}_{t-n}}{\text{Value}_{t-n}}$. The previous metrics evaluate the profits, but investors are often interested in risk management. We use the Sortino ratio based on the distribution of losses. It only penalizes downward volatility, defined as: 
$$\text{Sortino}=\frac{PR - RFR}{\sigma_d}, \sigma_d=\sqrt{\frac{1}{n}\sum_{<threshold}(PR-RFR)^2}$$
With PR, the annual returns, RFR, the risk-free rate, and $\sigma_d$, the distribution of downsides. For our backtest, the RFR and downside threshold are 0\%. The final metric is the Maximum Drawdown. It is defined as: 
$$MDD=\frac{\text{Bottom value} - \text{Peak value}}{\text{Peak value}}$$
It measures the maximum value loss an agent incurred during the period, where investors would generally discard algorithms with $MMD > 20\%$.

We define robustness as the ability of an agent to resist internal and external variations and uncertainty. Algorithms should be reliable in out-of-distribution data and volatile periods, which is expected to occur in unseen market conditions. We focus on observing the behavior of agents over time \cite{chan_measuring_2019}, in worse-case scenarios \cite{moos_robust_2022,subbaswamy_evaluating_2021}, and the relative performance between training and backtesting.

The first robustness metric is the Conditional Value at Risk. The literature defines it as: 
$$CVaR = \frac{1}{1-0.05} \int_{-1}^{VaR}xp(x)dx$$ 
Where VaR, the Value at Risk, is the average 5\%, a popular value, worse rate of returns, and x is a given rate lower than this threshold. CVaR measures the expected returns in worse-case scenarios. We use this metric because we expect DRL approaches to lack some of the stability that is found in traditional heuristic methods. Algorithms that manage returns with low kurtosis and, thus higher CVaR are deemed to be more stable.

The second robustness aspect to be evaluated is the generalization capabilities of the algorithms. We compare their performance during training and backtesting and the performance trend over sequential periods of the backtest. Both rely on the Information Ratio, a financial metric that describes an investment's performance beyond the market's returns. It is considered a measure of active management. 
$$IR = \frac{mean(PR_t - \text{IndexR}_{t})}{std(PR_t - \text{IndexR}_{t})})$$
Where $\text{IndexR}$ is the market's index returns. The first use-case is an IR Quotient, comparing the management performance in the validation set to the performance during the backtest: 
$$IR_q = \frac{IR_{\text{backtest}}}{IR_{\text{validation}}}$$
Values closer to 1 indicate the algorithm preserves its performance on out-of-distribution data. The second use case is the IR trend, evaluating the relevance of our trained algorithms with regard to market condition shifts. We compare the monthly performance of an agent to a static market index. The aim is to measure how well the learned behavior stays relevant over time. We define it as: 
$$IR_{trend} = \frac{\sum\limits_{t=0}^{T}(t-T/2)(IR_t-\bar{IR})}{\sum\limits_{t=0}^{T}(t-T/2)^2}$$
Both metrics rely on a market index. While our assets belong to the S\&P500, using its price as a point of comparison is unfair because we use an unrepresentative subset. Smaller portfolios are subject to more volatility, and hand-picking assets results in performance biases. The market index we use is defined as a Mean-Variance Optimization allocation at $t=0$ and rebalanced monthly during the period. This approach yields a more realistic base of comparison for our algorithms.

The third robustness aspect is the stability of the algorithms. We aim to distinguish one-off successes from actually learned behaviors. Using multiple seeds trained per agent, we evaluate the variance of metrics. Agents trained on the same data, with the same input, should yield similar results. Therefore, the lower the interseed variance, the more robust the approach.

\section{Results}

During the backtesting period, we gathered trajectories for 64 combinations of algorithms. Each metric is the average and standard deviation of performance over the 5 seeds. Due to the volume, we have chosen to represent the 10 combinations with the highest rates of returns, as well as the two worse, as a basis of comparison. The names are abbreviated versions of previous approaches, where "default" are daily prices, "lagged" are sparse windows, "net" are daily returns and "value" are value changes over an episode.

\begin{figure}
  \centering
  \includegraphics[width=0.5\textwidth]{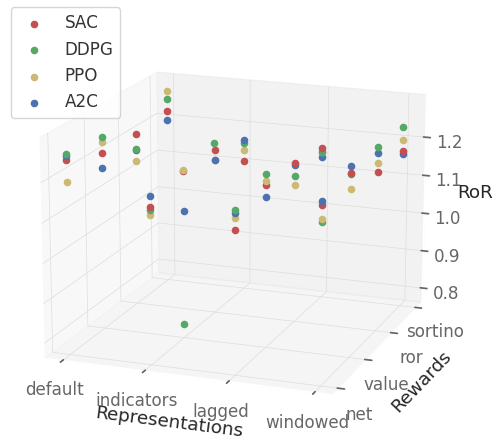}
  \caption{Rate of Return}
  \label{fig:value}
\end{figure}

\begin{table*}[htbp]\label{perf-metrics}
\caption{Performance Metrics}
\resizebox{\textwidth}{!}{%
\begin{tabular}{|c c c c c c c c c|} 
 \hline
 Rank & Name & RoR & Sortino & MDD & CVaR & IR & IR Trend & IR Quotient \\ [0.5ex] 
 \hline
  & MVO & 0.98 & -0.066 & -21.9\% & - & - & - & - \\
 \hline
 1 & ppo\_default\_sortino & \textbf{1.28} $\pm$ 0.05 & \textbf{0.94}  & \textbf{-18.9\%} & -2204 $\pm$ 71 & \textbf{0.05} & -0.0051 $\pm$ 0.002 & \textbf{0.76} $\pm$ 0.29 \\ 
 \hline
 2 & ddpg\_default\_net & 1.26 $\pm$ 0.06 & 0.82 & -20.9\% & -2418 $\pm$ 133 & 0.05 & -0.0071 $\pm$ 0.002 & 0.49 $\pm$ \textbf{0.04} \\
 \hline
 3 & a2c\_default\_net & 1.26 $\pm$ \textbf{0.02} & 0.88 & -19.2\% & -2214 $\pm$ \textbf{51} & 0.05 & -0.0068 $\pm$ \textbf{0.001} & 0.55 $\pm$ 0.06 \\
 \hline
 4 & ddpg\_default\_value & 1.25 $\pm$ 0.07 & 0.90 & -19.3\% & -2124 $\pm$ 52 & 0.05 & -0.0070 $\pm$ 0.003 & 0.59 $\pm$ 0.12 \\
 \hline
 5 & ddpg\_default\_sortino & 1.25 $\pm$ 0.05 & 0.89 & -19.4\% & -2123 $\pm$ 101 & 0.05 & -0.0047 $\pm$ 0.003 & 0.60 $\pm$ 0.17 \\
 \hline
 6 & sac\_default\_net & 1.25 $\pm$ 0.07 & 0.87 & -18.6\% & -2120 $\pm$ 121 & 0.05 & -0.0047 $\pm$ 0.001 & 0.62 $\pm$ 0.27 \\
 \hline
 7 & a2c\_windowed\_value & 1.24 $\pm$ 0.05 & 0.85 & -18.8\% & -2114 $\pm$ 207 & 0.04 & -0.0055 $\pm$ 0.002 & 0.56 $\pm$ 0.17 \\
 \hline
 8 & ppo\_default\_value & 1.24 $\pm$ 0.08 & 0.83 & -19.2\% & -2195 $\pm$ 86 & 0.04 & -0.0061 $\pm$ 0.003 & 0.51 $\pm$ 0.22 \\
 \hline
 9 & ddpg\_windowed\_sortino & 1.23 $\pm$ 0.06 & 0.85 & -19.2\% & -2071 $\pm$ 98 & 0.04 & \textbf{-0.0044} $\pm$ 0.001 & 0.53 $\pm$ 0.19 \\
 \hline
 10 & ddpg\_windowed\_ror & 1.23 $\pm$ 0.04 & 0.85 & -18.2\% & \textbf{-2067} $\pm$ 156 & 0.04 & -0.0045 $\pm$ 0.002 & 0.56 $\pm$ 0.09 \\
 \hline
 63 & a2c\_indicators\_value & 1.09 $\pm$ 0.10 & 0.31 & -24.7\% & -2064 $\pm$ 172 & 0.02 & -0.0065 $\pm$ 0.001 & 0.21 $\pm$ 0.29 \\
 \hline
 64 & ddpg\_indicators\_value & 0.80 $\pm$ 0.09 & -0.93 & -33.1\% & -1698 $\pm$ 185 & -0.04 & 0.0004 $\pm$ 0.004 & 0.68 $\pm$ 0.88 \\
 \hline
\end{tabular}%
}
\end{table*}

The results were obtained following an hyperparameter sweep, where we selected the best performing set. These sets of parameters were trained for enough iterations to reach a plateau and the best checkpoint was chosen. Each algorithm was trained to the best of our ability. From table~\ref{perf-metrics}, we can see that most top-performing approaches performed relatively close to each other in terms of performance, with net returns in the 23-28\% range over two years. Only the worst-performing approach lost wealth, on average, compared to the 63 others having positive returns. Most approaches reliably reached comparable results across seeds, with relatively low spread of performance. Most approaches beat out the Mean Variance Optimization algorithm by a fair margin. 

However, while most returns are interesting, their risk management is lacking, reaching high volatility with regard to the returns, as denoted by the Sortino ratio. This volatility is confirmed by the Maximum DrawDown measures, which are close to the limit of what may be tolerated for risky management.

From Fig. 1, we evaluate the effect of each component on algorithms' performance. The highest impact comes from the representations, where the daily prices performed best, followed closely by the continuous sliding window. Surprisingly, the indicators and sparse windows lead to markedly worse performance. The reward functions had no discernible impact on performance, with the average returns for each function being almost identical. However, RoR provided the best performance in most cases, with a lower variance in results, proving to be more reliable. Finally, the training algorithms had a low impact, each outperforming others for given combinations of states and rewards. DDPG came out better in most configurations, but did result in the single worse performance. We can conclude that an approach based on daily values, which aims to maximize episodic Rate of Returns using DDPG would generally yield the best results.

The robustness metrics indicate underlying problems for our algorithms. The IR Quotients for our agents lie around 0.6, indicating a large degradation of performance between the validation and backtesting performances. Agents learned to pick winning assets during the training phase, but these were no longer the best performing at later dates. This aspect is confirmed by the IR Trend, which indicates a monthly degradation in performance for the length of the backtest, as the market shifts. Both metrics point to overfitting of action policies, confirmed by analyzing allocations over a period, with no variations in positions for some agents. Generalizable behaviors should focus on recognizing patterns instead of remembering which assets were previously winning. We can point to the learning efficiency of the algorithms, which are not suited for the low volume of data available in OLPS.

\section{Conclusion}

We proposed a standardized training and evaluation process to evaluate the performance and robustness of a large scope of OLPS approaches. We found that most approaches performed relatively close to one another. Yet, the two types of metrics indicate opposing results. While the returns and risk management were interesting, fairly beating out a popular allocation algorithm, they do not generalize well to unseen data and their performance degrades markedly over time. This highlights the learning efficiency limits of popular algorithms applied to the OLPS problem. Further improvements to these aspects may yield more robust approaches that profitably manage portfolios for longer lengths of time. 

\printbibliography
\end{document}